\documentclass[letterpaper, 10 pt, conference]{ieeeconf}  

\IEEEoverridecommandlockouts                              

\title{\LARGE \bf
How Can One Choose the Best CAM-Based Explainability Method for a CNN Model?
}

\author{Daniel da Silva Costa$^{1}$, Pedro Nuno de Souza Moura$^{1}$ and Adriana C. F. Alvim$^{1}$
\thanks{$^{1}$Daniel da Silva Costa, Pedro Nuno de Souza Moura and Adriana C. F. Alvim are with Postgraduate Program in Informatics, Federal University of the State of Rio de Janeiro (UNIRIO), Rio de Janeiro, RJ, 22290-240, Brazil
{\tt\small daniel.scosta@edu.unirio.br, pedro.moura@uniriotec, adriana@uniriotec.br}} %
}
\usepackage{comment}

\usepackage{graphicx}

\usepackage{booktabs}
\usepackage{xcolor,colortbl}
\definecolor{lavender}{rgb}{0.9, 0.9, 0.98}
\definecolor{lavenderbold}{rgb}{0.65, 0.65, 0.9}
\usepackage{multirow}

\usepackage{array,booktabs}
\usepackage{lscape}
\usepackage{amsmath}  
\usepackage{amssymb}
\usepackage{url} 
\begin{document}

\maketitle
\thispagestyle{empty}
\pagestyle{empty}

\begin{abstract}

In recent years, several advances have been observed in Deep Learning with surprising results. Models in this area have been increasingly used in numerous applications, including those sensitive to human life, which require clear explanations and justifications. This has encouraged several types of research into the explainability of neural networks. 
Various explainability methods have been proposed, but not many metrics to evaluate these methods.
The most commonly used metric is the Intersection over Union (IoU). It is applied between two bounding boxes to allow one to measure the similarity between them. However, due to the characteristics of the results of the explainability methods, called saliency maps, which do not have a known shape, we hypothesise that there must be a better metric that allows one to find an explainability method that produces results that best resemble the human perception. 
This work proposes using different metrics to assess the similarity between human perception and the explanation saliency maps to find a better metric. To this end, an investigation was conducted employing a subset of the ImageNet dataset, which corresponded to the Chihuahuas images. 
Several CAM-based explainability methods were used to generate saliency maps, which were compared with human perception of the most important parts of the chihuahuas in each image. Alignment was measured by applying distance metrics between the bounding box of human annotations and the saliency maps produced by each explainability method. Rankings of the best saliency maps were created using the results of the distance metrics and compared to the ranking obtained using people's choice, collected through crowdsourcing, of the best explanation saliency maps for each selected image. Comparison between rankings was performed using the Rank-Biased Overlap (RBO) metric. The results indicate the feasibility of our method to find the explainability method that best resembles human perception. In our experiments, the two metrics that best resemble human perception corresponded to Manhattan and Correlation. Besides, the best explainability methods regarding human perception were LayerCAM, Score-CAM, and IS-CAM.

\end{abstract}

\section{INTRODUCTION}

In recent years, Artificial Intelligence models based on neural networks have become practically ubiquitous and have been used in a wide variety of contexts. This has been possible thanks to advances in Deep Learning (DL), which uses artificial neural networks (ANNs). These networks can internally build a complex model of the input data automatically and without the need for human operators to previously describe rules or characteristics of the data~\cite{goodfellow2016deep}.

Despite significant advances in DL and its recent popularization, some application areas require greater care and understanding of the information on which ANNs are based when learning internal representations of input data. Applications in more critical areas, such as medicine and law, usually require clear and unequivocal explanations and justifications for the systems' recommendations, as they deal with situations potentially harmful to human life. If the user does not understand the rules that generated the ANN model or decision process, the confidence in the results and the usefulness of these networks will be compromised~\cite{ras2022explainable}.

In this context, the area of Explainability (or Interpretability) is gaining more attention from members of society who need a more detailed understanding of what happens within DL models. Generally speaking, an explanation can be understood as any information, in any format, that can help the user understand why a given model has a specific decision pattern, either concerning the entire dataset or about an individual belonging to that set~\cite{ras2022explainable}. The scientific community has made efforts to promote the explainability of ANN models to increase the reliability of their use, facilitating knowledge about their internal rules and possibly helping to optimize and build better ANNs~\cite{ras2022explainable}.

In the case of the models trained for image classification problems in Computer Vision (CV), the most widely used explainability methods seek to show the user which information from the model's input is most relevant for that specific model for classifying a given image. To do this, they usually produce a saliency map (or heatmap) that is superimposed on the image to allow the decision process of these models to be understood.


Researchers commonly assess the quality of their results by depicting heatmaps (explanations) of their proposed method and other methods in the literature and visually comparing their highlighted areas with respect to the object of interest of the image.
Objective metrics have been proposed and are more common than human perception-based metrics; however, to the best of our knowledge, assessing the quality and reliability of these methods remains an open problem~\cite{colin2022cannot}.

We think a more confident way should be utilized to assess and compare the quality of the visual explanations, so as to allow a researcher or a user to choose the best explainability method for CV. We based our hypothesis on the idea that the best and most helpful explainability methods will be the ones that most resemble human perception~\cite{zhang2019dissonance, prasad2020extent}.


Recently, the scientific community has brought several papers to the discussion on evaluating explainability methods based more on human understanding and perception of human beings~\cite{colin2022cannot, kim2022hive, prasad2020extent}.

This work seeks to understand and evaluate the alignment between the results of explainability methods, which generate heatmaps, such as Class Activation Maps (CAM)~\cite{zhou2016learning} and its variations (``CAM-based''), such as Grad-CAM~\cite{selvaraju2017grad} and Score-CAM~\cite{wang2020score}, with people's perception of the most important area of each image for a specific classification. The word ``alignment'' can have various meanings depending on the area of knowledge employing it. 
In this work, the concept of alignment will be based on the similarity between human perception and the results generated by applying explainability methods~\cite{prasad2020extent}.

In CV, it is common to apply the metric Intersection over Union (IoU) (also called Jaccard similarity)~\cite{zhou2016learning} to calculate the similarity between two boxes. In our research, we started by using this metric to understand the alignment (similarity) between the bounding boxes of the human annotations of the object of interest in the image and the bounding box built using the heatmaps generated by explainability methods. 

\begin{figure*}[tb] 
  \centering
  \includegraphics[width=\textwidth]{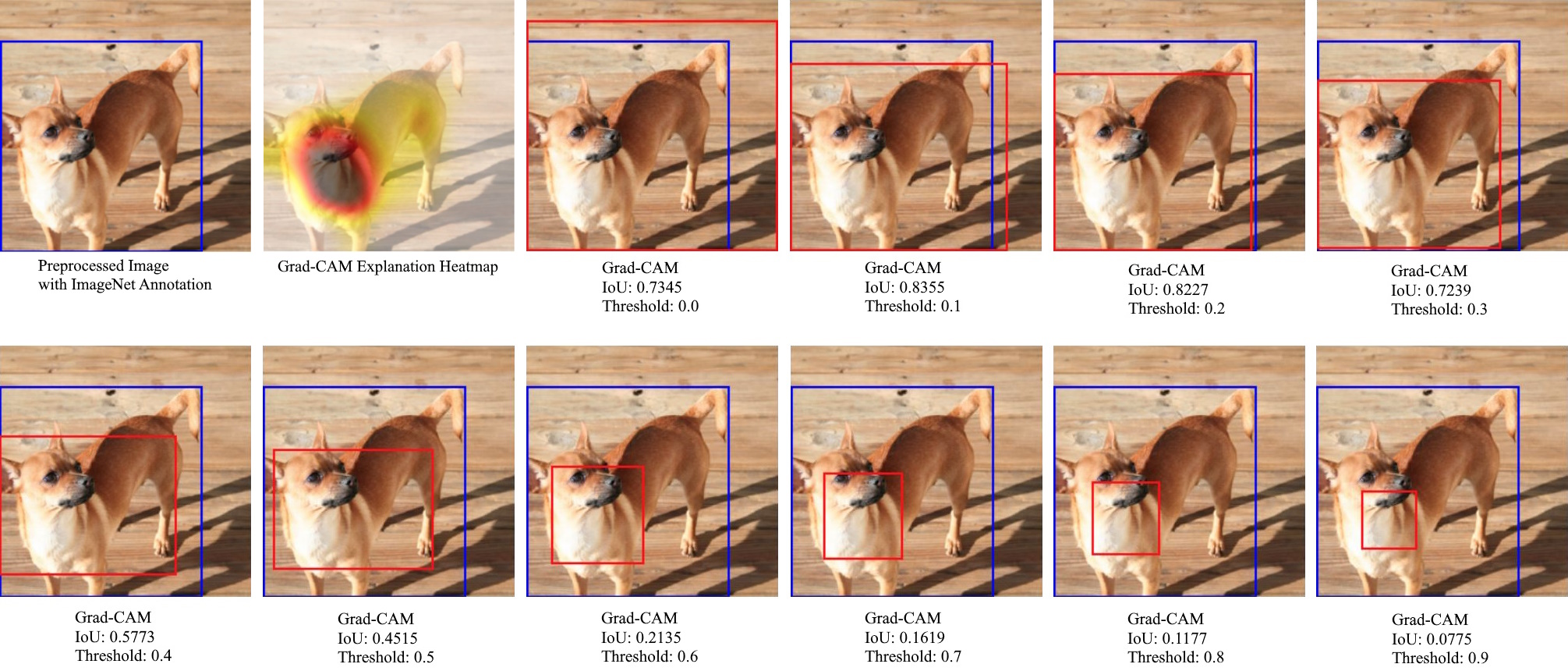}
  \caption{From the top-left corner: firstly, the preprocessed image n02085620\_5542 with the annotation bounding box, from ImageNet, is shown, in blue. Secondly, the Grad-CAM explanation heatmap. The darker the shade of red in the explanation, the more important the area. The lighter the yellow, the less important it is. The other images show in red the bounding boxes created by applying thresholds in the values of the Grad-CAM explanation heatmap.}
  \label{fig:n02085620_5542--thresholds--gradcam.jpg}
\end{figure*}

The problem with such an approach is determining the limit of the heatmap, since its area is not a box. Figure~\ref{fig:n02085620_5542--thresholds--gradcam.jpg} shows, from the top left corner, an image with the ImageNet annotation bounding box and secondly the Grad-CAM explanation heatmap. We tried to apply thresholds to its values, to remove values below each threshold. Figure~\ref{fig:n02085620_5542--thresholds--gradcam.jpg} shows the resulted bounding box in red and the value of IoU, below the image, for the Grad-CAM explainability method for the image n02085620\_5542. The blue bounding box is the ground truth annotation from the ImageNet dataset. It can be seen that the best results for this case were achieved using a threshold of $0.1$ (IoU: $0.8355$) and $0.2$ (IoU: $0.8227$). The problem is how to determine the best threshold since the results for each explainability method and image are not the same. Moreover, when applying a threshold, we lose information that can help to explain the decision of the model. Therefore, the present research ought to use different metrics that allow an investigation of the similarity between human perception and the heatmaps generated by the explainability methods, using all information of the heatmaps.

The results indicate the feasibility of our method to find the explainability method that best resembles human perception and suggest that: (i) the metrics that best resemble human perception are two: Manhattan and Correlation; and (ii) the best explainability methods regarding human perception were LayerCAM, Score-CAM, and IS-CAM.

The remainder of this paper is organised as follows: Section~\ref{sec:related-works} brings related work about the topic of this work. In Section~\ref{sec:methodology}, we describe our method, and in Section~\ref{sec:results-discussion}, we discuss the results. We present a brief resume and final considerations in Section~\ref{sec:conclusion}.

All codes and results of this paper are available in the GitHub\footnote{https://github.com/danieldasilvacosta/choose-best-cam-based-xai-method-cnn-model} repository built for this research.



\section{RELATED WORKS} \label{sec:related-works}


Methods based on CAM~\cite{zhou2016learning} have been widely used and are among the most traditional methods for assessing the importance given by the model to each pixel in an image~\cite{ras2022explainable, ibrahim2023explainable}. In general, the result of applying these methods is the information about which pixels are most important to the model in a specific image and for a specific class, which allows the user to visualise the areas of the image to which the model has given the most importance. The original CAM method allows the image to be visualised by constructing a heatmap of the discriminating areas used by a convolutional neural network (CNN)~\cite{lecun1989backpropagation} in the inference phase. CNNs correspond to the most traditional architecture used in CV, which perhaps justifies the number of explainability methods proposed for this architecture. In this work, a pre-trained neural network with an architecture based on CNN is used. 


In this work, we divide the explainability methods into two groups: activation-based and gradient-based methods. Gradient-based (or backpropagation-based~\cite{ras2022explainable}) methods calculate the importance of each input feature based on some evaluation of the gradients of the trained neural network that relate to each feature. The activation-based (or perturbation-based~\cite{ras2022explainable}) methods calculate the importance of each input feature based on the difference of the outputs of the trained neural network for an image and modified copies of that image.

Despite many proposals for CAM-based methods in recent years~\cite{zhou2016learning, wang2020score, wang2020ss, naidu2020cam, selvaraju2017grad, chattopadhay2018grad, omeiza2019smooth, fu2020axiom, jiang2021layercam}, to date, there have not been many papers dedicated to exploring and carrying out an exhaustive comparison of these methods to allow for further discussion in the scientific community.


A recent work similar to the one proposed in this paper is that of~\cite{morrison2023shared}. Although the authors carried out alignment experiments between human perception and the results of different explainability methods and network architectures, they did not take into account that alignment can have specific results depending on whether the explainability method is gradient-based or not. They only evaluated the results of gradient-based methods: Grad-CAM and XRAI~\cite{kapishnikov2019xrai}. The Score-CAM method, for example, is not based on gradients and has been one of the most cited in recent works. 

In this sense, a novelty presented by this work corresponds to the evaluation of the results of CAM-based explainability methods using human perception as ground truth, while considering both gradient-based and activation-based methods.



\section{METHODOLOGY} \label{sec:methodology}

The ImageNet~\cite{russakovsky2015ImageNet} dataset is used as a common benchmark in CV. It has a great volume and variety of images: there are $1,000$ classes and $1,281,167$ training images. Furthermore, this dataset has human annotations about the position of the target objects in the images, which are called bounding box annotations. An example of an image from the ImageNet dataset with its bounding box annotation can be seen in Figure~\ref{fig:n02085620_5542--thresholds--gradcam.jpg} on the top-left corner. Therefore, this dataset was a natural choice of dataset for our work.



We obtained all images from the Chihuahua class of the ImageNet dataset, filtering for those that only contain a single annotation bounding box, 
and randomly selecting $20$ images. This class has $1,300$ training images. This class was chosen because: (i) we considered that they are animals that people from different countries could easily recognize in the experiments; 
(ii) we wanted to see if the explainability methods had any problems with the characteristics of the dogs' fur and the background information of the image, since in several images both have similar colors, for example;
(iii) we considered that dogs have attributes that are similar and different from other living beings, such as their eyes and snout, and we would like to know if and how these attributes would be recognised by the model and the explainability methods.

We applied a pre-trained ResNet50 model~\cite{he2015resnet} to the selected images to obtain the classifications. This is a CNN traditionally used in the fields of CV and Explainability. 
We employed the model available in the PyTorch library with the following version of the weights: ImageNet1K\_V2\footnote{\url{https://pytorch.org/vision/main/models/generated/torchvision.models.resnet50.html}}. Each selected image was preprocessed using the ResNet50\_Weights.DEFAULT.weights.transforms() method in PyTorch. The final size of the preprocessed images was 224x224 pixels.  

One shall note that the model incorrectly classified $3$ images in this step, marked with an ``F'' in the ``Correct?'' column in Table~\ref{table:selected-images}. There were $17$ correctly classified images, marked with an ``T'' in the same column. The wrongly classified images were removed from the experiments and analysis of the results. The ``\% of Prediction'' column shows the likelihood of the chosen class indicated by the model. 

We then asked $50$ people in an experiment (``annotation experiment'') on the Appen crowdsourcing platform\footnote{https://www.appen.com/} to annotate the selected images by drawing a bounding box around the chihuahua in each of the $17$ selected images. 

\begin{figure}[tb]
  \centering
  \includegraphics[width=\columnwidth]{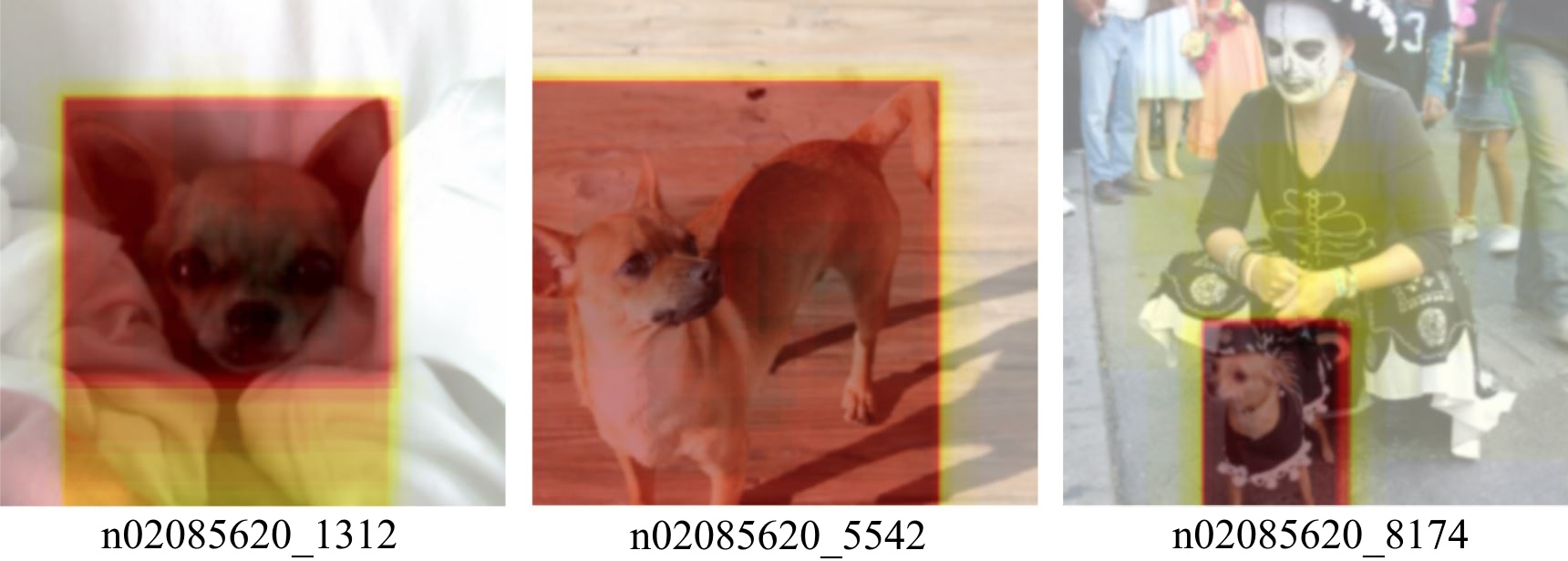}
  \caption{Examples of the final annotation heatmap for three images: n02085620\_1312, n02085620\_5542 and n02085620\_8174. The darker the shade of red, the more important the area. The lighter the yellow, the less important the area.}
  \label{fig:examples-appen-heatmap.jpg}
\end{figure}

With the aim of trying to better capture the importance of each pixel in the image w.r.t. the perception of people, we combined all the $50$ annotations in one ``annotation heatmap'' for each image. Each pixel in the final annotation heatmap is weighted w.r.t. its frequency in the $50$ annotations. For example, if a pixel is present in $5$ bounding boxes, it will have a weight of $5$ in the annotation heatmap. After computing the weight of each pixel, the values were normalised between $0$ and $1$. The Figure~\ref{fig:examples-appen-heatmap.jpg} depicts three examples of the final annotation heatmap that overlaps the preprocessed chihuahua image. The darker the shade of red, the more important the area. The lighter the yellow, the less important the area. For example, one can see that most of the annotations on the image n02085620\_1312 are over the face of the dog, but there is a selected area below the face, maybe because of the paw of the dog. In the n02085620\_8174 case, some participants highlighted parts of the human body beyond the dog.

\begin{table}[tb]
    \caption{The Selected Images from ImageNet and their ResNet50 Classification.}
    \label{table:selected-images}
    \begin{center}
        \begin{tabular}{>{\centering\arraybackslash}m{0.25\columnwidth}>{\centering\arraybackslash}m{0.25\columnwidth}>{\centering\arraybackslash}m{0.15\columnwidth}>{\centering\arraybackslash}m{0.15\columnwidth}}
        
            \toprule

            \textbf{Image} & \textbf{\shortstack{ResNet50 \\ Classification}} & \textbf{\shortstack{\% of \\ Prediction}} & \textbf{Correct?} \\

            \midrule
            
            \rowcolor{lavender}
            n02085620\_199 & Chihuahua  & 28.0\% & T \\
            
            n02085620\_242 & Chihuahua & 20.4\% & T \\
    
            \rowcolor{lavender}
            n02085620\_400 & \shortstack{soft-coated \\ wheaten terrier} & 6.6\% & F \\
            
            n02085620\_1312 & Chihuahua & 50.6\% & T \\
    
            \rowcolor{lavender}
            n02085620\_1558 & Chihuahua & 27.3\% & T \\
            
            n02085620\_1994 & Chihuahua & 46.1\% & T \\
    
            \rowcolor{lavender}
            n02085620\_2163 & Chihuahua & 39.4\% & T \\
            
            n02085620\_3078 & Chihuahua & 46.8\% & T \\
    
            \rowcolor{lavender}
            n02085620\_5002 & Chihuahua & 56.7\% & T \\
            
            n02085620\_5071 & Chihuahua & 38.7\% & T \\
    
            \rowcolor{lavender}
            n02085620\_5096 & Chihuahua & 44.0\% & T \\
            
            n02085620\_5542 & Chihuahua & 52.3\% & T \\
    
            \rowcolor{lavender}
            n02085620\_5629 & redbone & 11.1\% & F \\
            
            n02085620\_7234 & Chihuahua & 34.1\% & T \\
    
            \rowcolor{lavender}
            n02085620\_7243 & Chihuahua & 32.0\% & T \\
            
            n02085620\_7697 & Chihuahua & 23.7\% & T \\
    
            \rowcolor{lavender}
            n02085620\_8174 & Chihuahua & 43.3\% & T \\
            
            n02085620\_11143 & \shortstack{West Highland \\ white terrier} & 8.8\% & F \\
    
            \rowcolor{lavender}
            n02085620\_27480 & Chihuahua & 36.1\% & T \\
            
            n02085620\_44901 & Chihuahua & 47.8\% & T \\
            
            \bottomrule

        \end{tabular}
    \end{center}
\end{table}

At this point, CAM-based explainability methods were applied to the ResNet50 model to generate the ``explanation heatmaps''. The values of each pixel in each heatmap were normalised between $0$ and $1$. The employed CAM-based methods, as well as their source papers, are exhibited in Table~\ref{table:explainability-methods}, where the column ``Type'' indicates whether the method is gradient-based or activation-based. These methods correspond to those available in the TorchCAM\footnote{https://github.com/frgfm/torch-cam} library, which are generally used in the literature. 

\begin{table}[tb]
    \caption{The Selected Explainability Methods}
    \label{table:explainability-methods}
    \begin{center}
        \begin{tabular}{ccccc}

            \toprule
                
            \textbf{Method} & \textbf{Acronym} & \textbf{Type} & \textbf{Paper} & \textbf{Year} \\
            
            \midrule
            
            \rowcolor{lavender}
            CAM & CAM & Activation & 
            \cite{zhou2016learning} & 2016 \\
            
            SS-CAM & SSCAM & Activation & \cite{wang2020ss} & 2020 \\
            
            \rowcolor{lavender}
            IS-CAM & ISCAM & Activation & \cite{naidu2020cam} & 2020 \\
            
            Score-CAM & ScCAM & Activation & \cite{wang2020score} & 2020 \\
            
            \rowcolor{lavender}
            Grad-CAM & GCAM & Gradient & \cite{selvaraju2017grad} & 2017 \\
            
            Grad-CAM++ & GCAM++ & Gradient & \cite{chattopadhay2018grad} & 2018 \\
            
            \rowcolor{lavender}
            Smooth Grad-CAM++ & SGCAM++ & Gradient & \cite{omeiza2019smooth} & 2019 \\
            
            X-Grad-CAM & XGCAM & Gradient & \cite{fu2020axiom} & 2020 \\
            
            \rowcolor{lavender}
            LayerCAM & LCAM & Gradient & \cite{jiang2021layercam} & 2021 \\
            
            \bottomrule

        \end{tabular}
    \end{center}
\end{table}

\begin{figure*}[tb] 
  \centering
  \includegraphics[width=.8\textwidth]{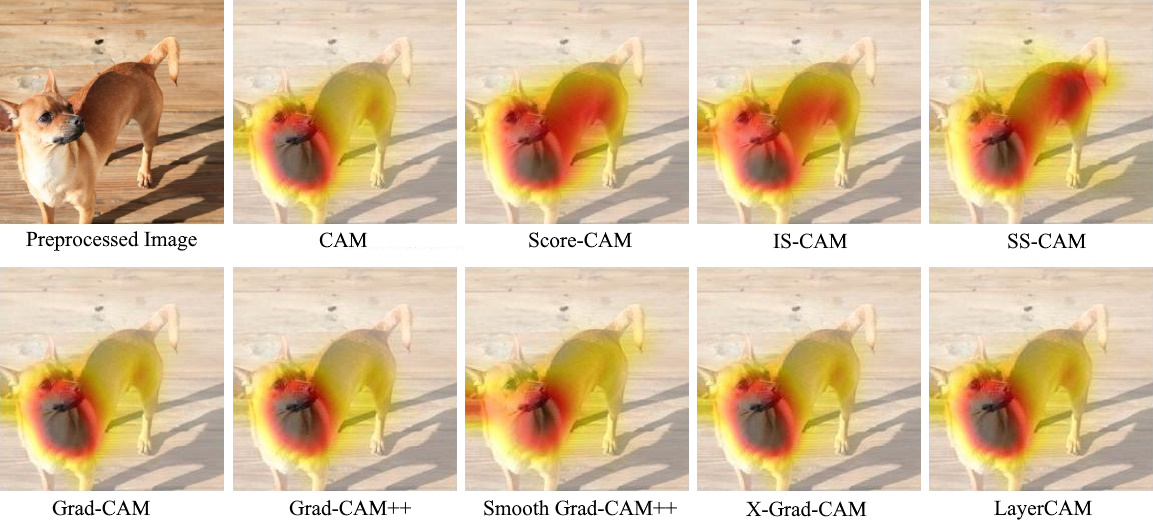}
  \caption{The preprocessed image n02085620\_5542 on the top-left and the overlapped explanation heatmap of each explainability method.}
  \label{fig:n02085620_5542--preprocessed--heatmaps.jpg}
\end{figure*}

Figure~\ref{fig:n02085620_5542--preprocessed--heatmaps.jpg} shows the image n02085620\_5542, after it was preprocessed and correctly classified by the ResNet50 model and the generated explanation heatmaps of the methods from Table~\ref{table:explainability-methods}. The darker the shade of red, the more important the area according to the method. The lighter the yellow, the less important the area. Although similar, the heatmaps are not the same, and one can notice, for example, the difference between the heatmaps from the methods Score-CAM and Smooth Grad-CAM++: the former gives more importance to different regions of the body of the dog than the latter.

We aim to find the explainability method that best aligns with human perception. To this end, we frame this problem as a problem of calculating the distance between the annotation heatmaps and the explanation heatmaps. The selected distance metrics were: (i) Weighted Jaccard (WJ); (ii) Wasserstein (also knows as Earth Mover's Distance) (WA); (iii) Bray-Curtis (BC); (iv) Canberra (CA); (v) Chebyshev (CY); (vi) Manhattan (MA); (vii) Correlation (CR); (viii) Cosine (CS); (ix) Euclidean (EU); (x) Jensen-Shannon (JS); (xi) Minkowski (MI); and (xii) squared Euclidean (SE). 




Each value in the explanation heatmaps represents the importance of each image pixel in the explanation. We used distance metrics to make it possible to capture these importances in order to obtain a fairer comparison between the explanation heatmaps and the annotation heatmaps.

We will not describe in detail all the distance metrics used in this research for the sake of space. We will only describe the Weighted Jaccard metric, which should allow the reader to understand how distance metrics can help to better capture the similarity between annotation and explanation heatmaps by considering the weight of each vector value in the calculations.


The Weighted Jaccard~\cite{ioffe2010improved} coefficient is a variation of the Jaccard coefficient (IoU), which allows distance measurement of two vectors based on the weights of the values at each position of each vector. The values should be real and positive numbers. It is easily calculated and, in our case, allows us to better grasp the importance of each pixel in the heatmaps. Equation (1) shows the Weighted Jaccard distance formula. 

$$
    d(u, v) = 1 - \frac{\sum_i min(u_i, v_i)}{\sum_i max(u_i, v_i)} \eqno{(1)}
$$

\noindent
where the vectors are represented by $u$ and $v$. The Weighted Jaccard distance generates values between $0$ and $1$, where the closer the result is to $0$, the closer the vectors are.


The Wasserstein distance between $u$ and $v$, is defined as:

$$
    d(u, v) = \inf_{\pi \in \Gamma (u, v)} \int_{R \times R} \left| x-y \right|_2 \mathrm{d} \pi (x, y) \eqno{(2)}
$$

\noindent
where $\Gamma (u, v)$ is the set of distributions on $\mathbb{R}^n \times \mathbb{R}^n$ whose marginals are $u$ and $v$ on the first and second factors, respectively. For a given value $x$, $u(x)$ gives the probability of $u$ at position $x$, and the same for $v(x)$.
The Bray-Curtis distance between $u$ and $v$ is defined as:

$$
    d(u,v) = \frac{\sum_i{|u_i-v_i|}}{\sum_i{|u_i+v_i|}} \eqno{(3)}
$$

The Canberra distance between $u$ and $v$ is defined as:

$$
    d(u,v) = \sum_i \frac{|u_i-v_i|}{|u_i|+|v_i|} \eqno{(4)}
$$

The Chebyshev distance between $u$ and $v$ is defined as:

$$
   d(u,v) = \max_{1 \le i \le n} |u_i-v_i| \eqno{(5)}
$$

The Manhattan distance between $u$ and $v$ is defined as:

$$
    d(u,v) = \sum_i {\left| u_i - v_i \right|} \eqno{(6)}
$$

The Correlation distance between $u$ and $v$ is defined as:

$$
    d(u,v) = 1 - \frac{(u - \bar{u}) \cdot (v - \bar{v})}
                   {{\|(u - \bar{u})\|}_2 {\|(v - \bar{v})\|}_2} \eqno{(7)}
$$

The Cosine distance between $u$ and $v$ is defined as:

$$
    d(u,v) = 1 - \frac{u \cdot v}
                   {{\|u\|}_2 {\|v\|}_2} \eqno{(8)}
$$

The Euclidean distance between $u$ and $v$ is defined as:

$$
    d(u,v) = \left(\sum_i{(u_i - v_i)^2}\right)^{1/2} \eqno{(9)}
$$

The Jensen-Shannon distance between $u$ and $v$ is defined as:

$$
    d(u,v) = \sqrt{\frac{D(u \parallel m) + D(v \parallel m)}{2}} \eqno{(10)}
$$

\noindent
where $m$ is the pointwise mean of $u$ and $v$ and $D$ is the Kullback-Leibler divergence.

The Minkowski distance between $u$ and $v$ is defined as:

$$
    d(u,v) = {\|u-v\|}_p = (\sum{|u_i - v_i|^p})^{1/p} \eqno{(11)}
$$

\noindent
where $p$ is the norm of the difference ${\|u-v\|}$. We applied $p = 3$.

The squared Euclidean distance between $u$ and $v$ is defined as:

$$
    d(u,v) = \sum_i{(u_i - v_i)^2} \eqno{(12)}
$$


The chihuahua images with the explanation heatmaps were shown to another group of people, in a different experiment (``validation experiment''), running on the Prolific crowdsourcing platform\footnote{https://www.prolific.com/}. For each selected image, we showed it overlapped by each explanation heatmap generated by the employed methods, similar to Fig.~\ref{fig:n02085620_5542--preprocessed--heatmaps.jpg} but without naming the methods. These superimposed images were randomly displayed to each different person in such a way that the explanation heatmaps generated by the same method appeared in different positions for each chihuahua image and for each participant. We asked participants to choose the option (explanation heatmap) that best explained the most important parts of the chihuahua in each image, in their opinion. Each participant had to choose one and only one option per chihuahua image. We used those choices to build a human ranking of the best explainability methods. We also built rankings based on the results of each distance metric (``metric ranking''). In this way, we could compare the human ranking with each metric ranking to find the distance metric that best resembles the human perception.


To compare the rankings, we used the Rank-Biased Overlap (RBO) metric, which is a similarity measure for indefinite rankings~\cite{webber2010similarity}. It was proposed to have specific qualities: (i) handle non-conjointness; (ii) weight high rankings more heavily than low; (iii) and be monotonic with increasing depth of evaluation. In particular, for this work, we wanted a metric that would allow us to evaluate the similarity between two rankings that considered more the top positions in each list and also allowed us to compare two lists of different sizes, given that not all explainability methods received votes during the validation experiment.

The RBO similarity formula, as defined in~\cite{webber2010similarity}, is described in Equation (13). In our case, we used the distance: $1$ - $\text{RBO}_{similarity}$. 
$$
    \text{RBO}_{similarity}(S, T, p) = (1-p) \sum_{d=1}^{\infty} p^{d-1} \cdot A_d \eqno{(13)}
$$

\noindent
where $S$ and $T$ are two infinite rankings; $p$ is the persistence parameter and indicates how the weights should be distributed throughout the ranking: to give more importance to the top-positions elements of the ranking, the smaller $p$ should be used; $d$ is the depth of the smaller list; and $A_d$ is the agreement of the lists with regard to $d$.

\section{RESULTS AND DISCUSSION} \label{sec:results-discussion}

In this section, we present and discuss the results obtained in our work. Table~\ref{table:n02085620_5542-metrics-values} exhibits the scores for each distance metric calculated between the annotation heatmap and each explanation heatmap for the image n02085620\_5542. We did the same for all the $17$ images. The lower the score, the better. The best result for each metric was highlighted in bold. In general, for this image, the activation-based methods were better. It is important to note that, in general, activation-based methods spent much more time to generate the heatmaps for all the $17$ images: GCAM (12.7 s), GCAM++ (13 s), SGCAM++ (34.2 s), XGCAM (14 s), LCAM (14.9 s), CAM (10.7 s), ScCAM (51min 47s), SSCAM (1d 5h 8min 21s), ISCAM (8h 5min 26s).


SSCAM obtained the best result in $8$ metrics, followed by ISCAM ($3$) and ScCAM ($2$). The Chebyshev metric did not help select the best method, as the best value, $0.0000$, appeared for several methods. It is also possible to see that the XGCAM scores were the same as the GCAM scores for all metrics. This was observed for all $17$ chihuahua images. We were unable to identify the cause of this, but perhaps it is due to a bug in the TorchCAM library.

\begin{table*}[tb]
    \caption{Calculated distance normalized values between the annotation heatmaps and the explanation heatmaps, for the image n02085620\_5542, for all 9 explainability methods, and for all distance metrics. The lower the score, the better.}
    \label{table:n02085620_5542-metrics-values}
    \begin{center}
        \begin{tabular}{cccccccccc}

            \toprule

            & \multicolumn{9}{c}{\textbf{Explainability Methods}} \\

            \textbf{\shortstack{Distance \\ Metrics}}
            & CAM 
            & SSCAM 
            & ISCAM 
            & ScCAM 
            & GCAM 
            & GCAM++ 
            & SGCAM++ 
            & XGCAM 
            & LCAM \\

            \midrule
            
            \rowcolor{lavender}
            WJ & 0.6331 & \textbf{0.0000} & 0.1291 & 0.1186 & 0.7394 & 1.0000 & 0.5915 & 0.7394 & 0.6478 \\

            WA & 0.6197 & \textbf{0.0000} & 0.2219 & 0.1947 & 0.7477 & 1.0000 & 0.5777 & 0.7477 & 0.6494 \\

            \rowcolor{lavender}
            BC & 0.6129 & \textbf{0.0000} & 0.1198 & 0.1099 & 0.7225 & 1.0000 & 0.5707 & 0.7225 & 0.6280 \\
            
            CA & 1.0000 & 0.4674 & \textbf{0.0000} & 0.0053 & 0.1518 & 0.1911 & 0.7140 & 0.1518 & 0.7374 \\

            \rowcolor{lavender}
            CY & \textbf{0.0000} & 0.3333 & \textbf{0.0000} & \textbf{0.0000} & \textbf{0.0000} & 1.0000 & \textbf{0.0000} & \textbf{0.0000} & \textbf{0.0000} \\ 
            
            MA & 0.6513 & 0.1041 & 0.0024 & \textbf{0.0000} & 0.7259 & 1.0000 & 0.5922 & 0.7259 & 0.6330 \\ 
            
            \rowcolor{lavender}
            CR & 0.8450 & 0.7031 & \textbf{0.0000} & 0.1530 & 0.7889 & 1.0000 & 0.5291 & 0.7889 & 0.7351 \\ 
            
            CS & 0.6289 & \textbf{0.0000} & 0.2410 & 0.3332 & 0.7582 & 1.0000 & 0.3428 & 0.7582 & 0.6684 \\ 
            
            \rowcolor{lavender}
            EU & 0.6648 & \textbf{0.0000} & 0.2623 & 0.2966 & 0.7767 & 1.0000 & 0.5565 & 0.7767 & 0.6952 \\ 
            
            JS & 0.5859 & \textbf{0.0000} & 0.1576 & 0.2927 & 0.7162 & 1.0000 & 0.2342 & 0.7162 & 0.6455 \\ 
            
            \rowcolor{lavender}
            MI & 0.5370 & \textbf{0.0000} & 0.8306 & 0.6830 & 0.7437 & 1.0000 & 0.4393 & 0.7437 & 0.6185 \\ 
            
            SE & 0.6438 & \textbf{0.0000} & 0.2441 & 0.2771 & 0.7604 & 1.0000 & 0.5334 & 0.7604 & 0.6753 \\

            \bottomrule
            
        \end{tabular}
    \end{center}
\end{table*}

\begin{figure}[tb]
  \centering
  \includegraphics[width=\columnwidth]{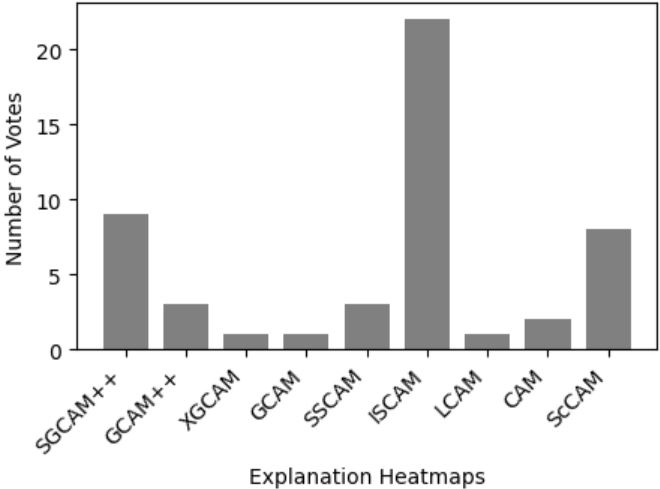}
  \caption{Selected explainability methods in the validation experiment for the image n02085620\_5542.}
  \label{fig:selected-heatmaps-bar-graph-n02085620_5542.JPG}
\end{figure}

Figure~\ref{fig:selected-heatmaps-bar-graph-n02085620_5542.JPG} shows the total of votes each method received for the image n02085620\_5542 in the validation experiment. In this case, the ISCAM method was the best regarding human perception, obtaining $22$ votes out of a total of 50. 

Table~\ref{table:total-votes-explainability-method} shows the number of votes for each method obtained as the main explanation for each image (column ``\# 1st''). The methods that have been voted first the most times 
were ISCAM, ScCAM and LCAM, whose values are highlighted in the table. Taking into account all $850$ possible votes in the validation experiment, which is the number of participants ($50$) multiplied by the number of images ($17$), it can be seen that ScCAM is the most popular with $174$ votes.

It is important to note that the GCAM method, although widely used in the literature, obtained the worst result concerning human perception with $54$ votes.

\begin{table}[tb]
    \caption{Total of votes of each explainability method.}
    \label{table:total-votes-explainability-method}
    \begin{center}
        \begin{tabular}{ccc}

            \toprule
                
            \textbf{Method} & \textbf{\# 1st} & \textbf{Total of Votes} \\
            
            \midrule
            
            \rowcolor{lavender}
            CAM & 0 & 68\\
            
            SSCAM & 3 & 129\\
            
            \rowcolor{lavender}
            ISCAM & \textbf{4} & 128\\
            
            ScCAM & \textbf{4} & \textbf{174}\\
            
            \rowcolor{lavender}
            GCAM & 0 & 54\\
            
            GCAM++ & 1 & 70\\
            
            \rowcolor{lavender}
            SGCAM++ & 1 & 85\\
            
            XGCAM & 0 & 55\\
            
            \rowcolor{lavender}
            LCAM & \textbf{4} & 87\\
            
            \bottomrule

        \end{tabular}
    \end{center}
\end{table}

Table~\ref{table:rankings-explainability-methods-image-n02085620_1312} shows the ranking built using the results of the human selection and distance metrics on the best explanation heatmaps for the image n02085620\_1312. We did the same for all the $17$ images. The column ``Rankings'' lists the name of the source of the created ranking: ``H'' for the human list and the rest for each distance metric. The columns ``1st'', ``2nd'', ``3rd'' until ``9th'' are the positions of the importance of each ranking, in such a way that ``1st'' is the most important position in that ranking and ``9th'' the least important one. The column ``RBO'' is the score achieved by each metric ranking compared to the human ranking, using the parameter $p$ equal to $1.0$. The closer the RBO score is to $0$, the closer that ranking is to the human ranking. Conversely, the closer it is to $1$, the further away. Therefore, it is possible to see that the Manhattan (MA) ranking, in bold, is the most similar to the human ranking, with an RBO score equal to $0.3347$ for the image n02085620\_1312.

\begin{table*}[tb]
    \caption{Rankings of explainability methods for the image n02085620\_1312.}
    \label{table:rankings-explainability-methods-image-n02085620_1312}
    \begin{center}
        \begin{tabular}{ccccccccccc}

            \toprule
                
            \textbf{Rankings} & \textbf{1st} & \textbf{2nd} & \textbf{3rd} & \textbf{4th} & \textbf{5th} & \textbf{6th} & \textbf{7th} & \textbf{8th} & \textbf{9th} & \textbf{\shortstack{RBO}} \\
            
            \midrule

            \rowcolor{lavenderbold}
            \textbf{H} & \textbf{LCAM} & \textbf{CAM} & \textbf{XGCAM} & \textbf{ScCAM} & \textbf{GCAM} & \textbf{GCAM++} & \textbf{ISCAM} & \textbf{-} & \textbf{-} & 0.0000 \\

            \midrule
            
            \rowcolor{lavender}
            WJ & SSCAM & SGCAM++ & CAM & LCAM & GCAM & XGCAM & ScCAM & ISCAM & GCAM++ & 0.5980 \\
            
            WA & SSCAM & SGCAM++ & LCAM & CAM & GCAM & XGCAM & ScCAM & ISCAM & GCAM++ & 0.5980 \\
            
            \rowcolor{lavender}
            BC & SSCAM & SGCAM++ & CAM & LCAM & GCAM & XGCAM & ScCAM & ISCAM & GCAM++ & 0.5980 \\

            CA & ISCAM & GCAM++ & ScCAM & GCAM & XGCAM & SSCAM & SGCAM++ & LCAM & CAM & 0.6813 \\
            
            \rowcolor{lavender}
            CY & SSCAM & CAM & LCAM & ScCAM & ISCAM & SGCAM++ & GCAM & GCAM++ & XGCAM & 0.4670 \\
            
            \textbf{MA} & \textbf{CAM} & \textbf{LCAM} & \textbf{SGCAM++} & \textbf{GCAM} & \textbf{XGCAM} & \textbf{ScCAM} & \textbf{ISCAM} & \textbf{GCAM++} & \textbf{SSCAM} & \textbf{0.3347} \\
            
            \rowcolor{lavender}
            CR & ScCAM & LCAM & CAM & ISCAM & SGCAM++ & GCAM & XGCAM & GCAM++ & SSCAM & 0.4228 \\
            
            CS & SGCAM++ & SSCAM & CAM & LCAM & ScCAM & GCAM & XGCAM & ISCAM & GCAM++ & 0.5980 \\
            
            \rowcolor{lavender}
            EU & SSCAM & SGCAM++ & CAM & LCAM & ScCAM & GCAM & XGCAM & ISCAM & GCAM++ & 0.5980 \\

            JS & SGCAM++ & CAM & LCAM & ScCAM & SSCAM & GCAM & XGCAM & ISCAM & GCAM++ & 0.4432 \\

            \rowcolor{lavender}
            MI & SSCAM & SGCAM++ & CAM & LCAM & ScCAM & GCAM & XGCAM & ISCAM & GCAM++ & 0.5980 \\

            SE & SSCAM & SGCAM++ & CAM & LCAM & ScCAM & GCAM & XGCAM & ISCAM & GCAM++ & 0.5980 \\

            \midrule

            \multicolumn{10}{c}{The best metric according to RBO distance ($p$ = 1.0): \textbf{Manhattan (MA)}.} \\
            
            \bottomrule

        \end{tabular}
    \end{center}
\end{table*}

Table~\ref{table:best-metric-rbo} shows the number of times each metric ranking was the most similar to the human ranking when the RBO metric was applied. The header columns $p$ are the values chosen for the parameter $p$ in the RBO. For each $p$ column in the table, it shows the total of images where that metric obtained the best RBO. The best values for RBO in each $p$ column are highlighted in bold. One shall note that, for some images, there were two or more rankings with the same score result. Therefore, if someone adds all values in each ``p'' column, the values will not necessarily add up to $17$. 

The parameter $p$ corresponds to a probability and is used to determine the weight of the sequence of items in each ranking. The weights are distributed in geometric progression; therefore, the sum of the weights is equal to $1$. The lower the value of $p$, the greater the importance attributed to the first items in the ranking, and the higher the value of $p$, the more evenly the importance is distributed among the other items, as can be seen in Table~\ref{table:weights-first-3-positions-rbo}. The columns ``First Position'', ``Second Position'', and ``Third Position'', indicating the ranking position. The percentage values are the importance of each position in the formula. We think that the analysis of the best result should consider more the importance of the first $3$ positions of the rankings, since the RBO for $p=0.0$ only considered the first position of the rankings, as can be seen in Table~\ref{table:weights-first-3-positions-rbo}. Therefore, we think that we should take into consideration the results for some value of $p$ between $p=0.5$ and $p=0.8$. In this case, looking at Table~\ref{table:best-metric-rbo}, the two best metrics were Manhattan (MA) and Correlation (CR), which were both $6$ times indicated as the most similar to the human ranking for RBO with $p=0.5$ and $p=0.8$. The ranking of explainability methods built with the Chebyshev (CY) metric obtained the worst result.

\begin{table}[tb]
    \caption{The total of images where each metric obtained the best RBO score for each selected $p$ value.}
    \label{table:best-metric-rbo}
    \begin{center}
        \begin{tabular}{cccccc}

            \toprule
                
            \textbf{Metric Ranking} & \textbf{p=0.0} & \textbf{p=0.5} & \textbf{p=0.8} & \textbf{p=0.9} & \textbf{p=1.0} \\
            
            \midrule

            \rowcolor{lavender}
            \textbf{WJ} & 7 & 4 & 4 & 2 & 3 \\ 
            
            \textbf{WA} & 5 & 5 & 5 & 3 & 4 \\ 

            \rowcolor{lavender}
            \textbf{BC} & 7 & 4 & 4 & 2 & 3 \\ 
            
            \textbf{CA} & 4 & 3 & 2 & 1 & 2 \\ 
            
            \rowcolor{lavender}
            \textbf{CY} & 1 & 0 & 0 & 1 & 1 \\ 
            
            \textbf{MA} & 6 & \textbf{6} & \textbf{6} & 4 & 5 \\ 
            
            \rowcolor{lavender}
            \textbf{CR} & \textbf{8} & \textbf{6} & \textbf{6} & \textbf{6} & 5 \\ 
            
            \textbf{CS} & 6 & 4 & 4 & 4 & \textbf{6} \\ 
            
            \rowcolor{lavender}
            \textbf{EU} & 6 & 4 & 4 & 2 & 3 \\ 
            
            \textbf{JS} & 4 & 2 & 2 & 2 & 4 \\ 
            
            \rowcolor{lavender}
            \textbf{MI} & 6 & 4 & 5 & 4 & 4 \\ 
            
            \textbf{SE} & 6 & 4 & 4 & 2 & 3 \\

            \bottomrule

        \end{tabular}
    \end{center}
\end{table}

\begin{table}[tb]
    \caption{The weights for each of the first $3$ positions in a ranking according to the $p$ value in the RBO metric.}
    \label{table:weights-first-3-positions-rbo}
    \begin{center}
        \begin{tabular}{cccc}

            \toprule
            
            \textbf{p} & \textbf{First Position} & \textbf{Second Position} & \textbf{Third Position} \\
            
            \midrule

            \rowcolor{lavender}
            0.0 & 100.0\% & 0.0\% & 0.0\% \\ 
            
            0.5 & 50\% & 25\% & 12.5\% \\ 
            
            \rowcolor{lavender}
            0.8 & 20\% & 16\% & 12.8\% \\ 

            0.9 & 10\% & 9\% & 8.1\% \\
            
            \rowcolor{lavender}
            1.0 & 0.0\% & 0.0\% & 0.0\% \\
            
            \bottomrule

        \end{tabular}
    \end{center}
\end{table}

\section{CONCLUSION} \label{sec:conclusion}

In this research, our objective was to find a better way to choose the best explainability method. We hypothesize that the best explainability method should resemble human perception as much as possible. To this end, we proposed a method in which we applied different distance metrics between the annotation heatmap created on several bounding boxes collected in an experiment carried out and the explanation heatmaps generated with the CAM-based explainability methods. With the results of the distance metrics, we conducted another experiment, through crowdsourcing, to ask participants to choose the best explanations and built a ranking of the choices to compare with the rankings created using the distance metrics. We did this comparison using the RBO metric. The obtained results suggest that: (i) the metrics that best resemble the human perception are two: Manhattan and Correlation; and (ii) the best explainability methods regarding human perception were LayerCAM, Score-CAM, and IS-CAM, for $4$ images each, and SS-CAM, for $3$, out of the total of $17$ chihuahuas images selected from the ImageNet dataset. This is an interesting result, as three of these methods are of the same type: activation-based. Unfortunately, these methods have the drawback 
of being too slow to be executed, requiring a lot of computational resources. One shall note that popular explainability methods, such as Grad-CAM, did not achieve the best results according to our method. 

For future work, we intend to apply our method for different settings in order to generalize the obtained results, such as: (i) using different CNN architectures; (ii) collecting annotations in polygon shapes, instead of the used rectangular shapes; (iii) using annotations created by agents based on Vision Language Models; and (iv) using images from different ImageNet classes or other datasets, considering a larger number of images.





\section*{ACKNOWLEDGMENT}

The authors would like to thank the Brazilian funding agency Coordenação de Aperfeiçoamento de Pessoal de Nível Superior (CAPES), Finance Code 001, for the partial financial support of this research.

\bibliographystyle{IEEEtran}
\bibliography{references.bib}

\end{document}